\documentclass{article}

\usepackage{arxiv}

\usepackage[utf8]{inputenc} 
\usepackage[T1]{fontenc}    
\usepackage{url}            
\usepackage{booktabs}       
\usepackage{amsfonts}       
\usepackage{nicefrac}       
\usepackage{microtype}      
\usepackage{lipsum}
\usepackage{graphicx}
\usepackage{multirow}
\usepackage{booktabs}
\usepackage[table,xcdraw]{xcolor}
\usepackage[hidelinks]{hyperref}
\graphicspath{ {./images/} }

\title{Toward Clinically Trustworthy Deep Learning}

\author{
\href{https://orcid.org/0009-0009-5139-4875}{\includegraphics[scale=0.06]{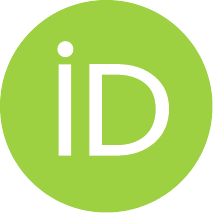}\hspace{1mm}Cooper Gamble$^+$} \\
  Radiology Informatics Lab\\
  Mayo Clinic\\
  Rochester, MN 55904 \\
  \texttt{Gamble.Cooper@mayo.edu} \\
   \And
   \href{https://orcid.org/0000-0003-3275-2971}{\includegraphics[scale=0.06]{orcid.pdf}\hspace{1mm}Shahriar Faghani$^+$} \\
  Radiology Informatics Lab\\
  Mayo Clinic\\
  Rochester, MN 55904 \\
  \texttt{Faghani.Shahriar@mayo.edu} \\
  \AND 
  \href{https://orcid.org/0000-0001-7926-6095}{\includegraphics[scale=0.06]{orcid.pdf}\hspace{1mm}Bradley J. Erickson$^*$} \\
  Radiology Informatics Lab\\
  Mayo Clinic\\
  Rochester, MN 55904 \\
  \texttt{bje@mayo.edu} \\
}
\renewcommand{\headeright}{}
\renewcommand{\undertitle}{Applying Conformal Prediction to Intracranial Hemorrhage Detection}
\renewcommand{\shorttitle}{}

\begin{document}
\pagestyle{plain}
\maketitle
\vspace{-20pt}
\begin{center}
	\footnotesize{$^+$Co-first author, $^*$Corresponding author} \\
\end{center}
\vspace{8pt}
\begin{abstract}
	As deep learning (DL) continues to demonstrate its ability in radiological tasks, it is critical that we optimize clinical DL solutions to include safety. One of the principal concerns in the clinical adoption of DL tools is trust. This study aims to apply conformal prediction as a step toward trustworthiness for DL in radiology. This is a retrospective study of 491 non-contrast head CTs from the CQ500 dataset, in which three senior radiologists annotated slices containing intracranial hemorrhage (ICH). The dataset was split into definite and challenging subsets, where challenging images were defined to those in which there was disagreement among readers. A DL model was trained on 146 patients (10,815 slices) from the definite data (training dataset) to perform ICH localization and classification for five classes of ICH. To develop an uncertainty-aware DL model, 1,546 cases of the definite data (calibration dataset) was used for Mondrian conformal prediction (MCP). The uncertainty-aware DL model was tested on 8,401 definite and challenging cases to assess its ability to identify challenging cases. After the MCP procedure, the model achieved an F1 score of 0.920 for ICH classification on the test dataset. Additionally, it correctly identified 6,837 of the 6,856 total challenging cases as challenging (99.7\% accuracy). It did not incorrectly label any definite cases as challenging. The uncertainty-aware ICH detector performs on par with state-of-the-art models. MCP’s performance in detecting challenging cases demonstrates that it is useful in automated ICH detection and promising for trustworthiness in radiological DL. 
\end{abstract}


\section{Introduction}
Intracranial hemorrhage (ICH) is a widespread and potentially life-threatening condition, affecting more than 50,000 adults annually in the United States (1). Deep learning (DL) has demonstrated promising results as an automated solution to ICH detection (2). However, because DL generally do not provide a confidence score for their predictions, it is difficult for both clinicians and researchers to trust them (3).
 
In general, a radiological DL model provides some predictions (diagnoses) in the form of numerical outputs for the possible diagnoses that were included in the training dataset. These outputs possess the mathematical properties of measures of probability, but they are not representative of the true probability of a given outcome because they are uncalibrated (3,4). Calibration is the process by which a model’s predictions are made to more accurately approximate the underlying distribution of the population on which it will be applied. Conventionally, calibration maps a model’s output to the observed frequency of an event. In this study, however, calibration refers to the process of learning the test distribution from a small sample of the population called the calibration dataset. Conformal prediction (i.e. distribution-free uncertainty quantification) is a family of statistical methods, the premise of which is to generate prediction sets (i.e., differential diagnoses) which are guaranteed to contain the true value (diagnosis) at a user-specified confidence level (5). Though conformal methods vary in nature and implementation, the general process begins by defining a heuristic notion of uncertainty (HNU) for the DL model. Then, for every instance in the calibration dataset, this uncertainty metric is calculated and recorded. Finally, the uncertainty values obtained during calibration are then applied to some computation during inference to yield sets of predictions—instead of point predictions—for each test instance. The nature of this computation depends on the nature of the selected conformal method.
 
Mondrian conformal prediction (MCP) is a subset of conformal prediction which relies on a grouping function to split the calibration data based on some shared characteristic (6). MCP operates by ranking the calibration HNU values across a given group, which allows for the calculation of conformal scores for predictions on the test dataset. For each test sample, a conformal score is obtained for each group by dividing the would-be rank (i.e., the insertion index) of the sample’s HNU value within a given group by the size of that group. This conformal score is then compared against a user-specified threshold, and groups for which the sample’s score exceeds the threshold are included in its final prediction set (see Figure 1).
\begin{figure}[h!]
	\centering
	\includegraphics[scale=0.35]{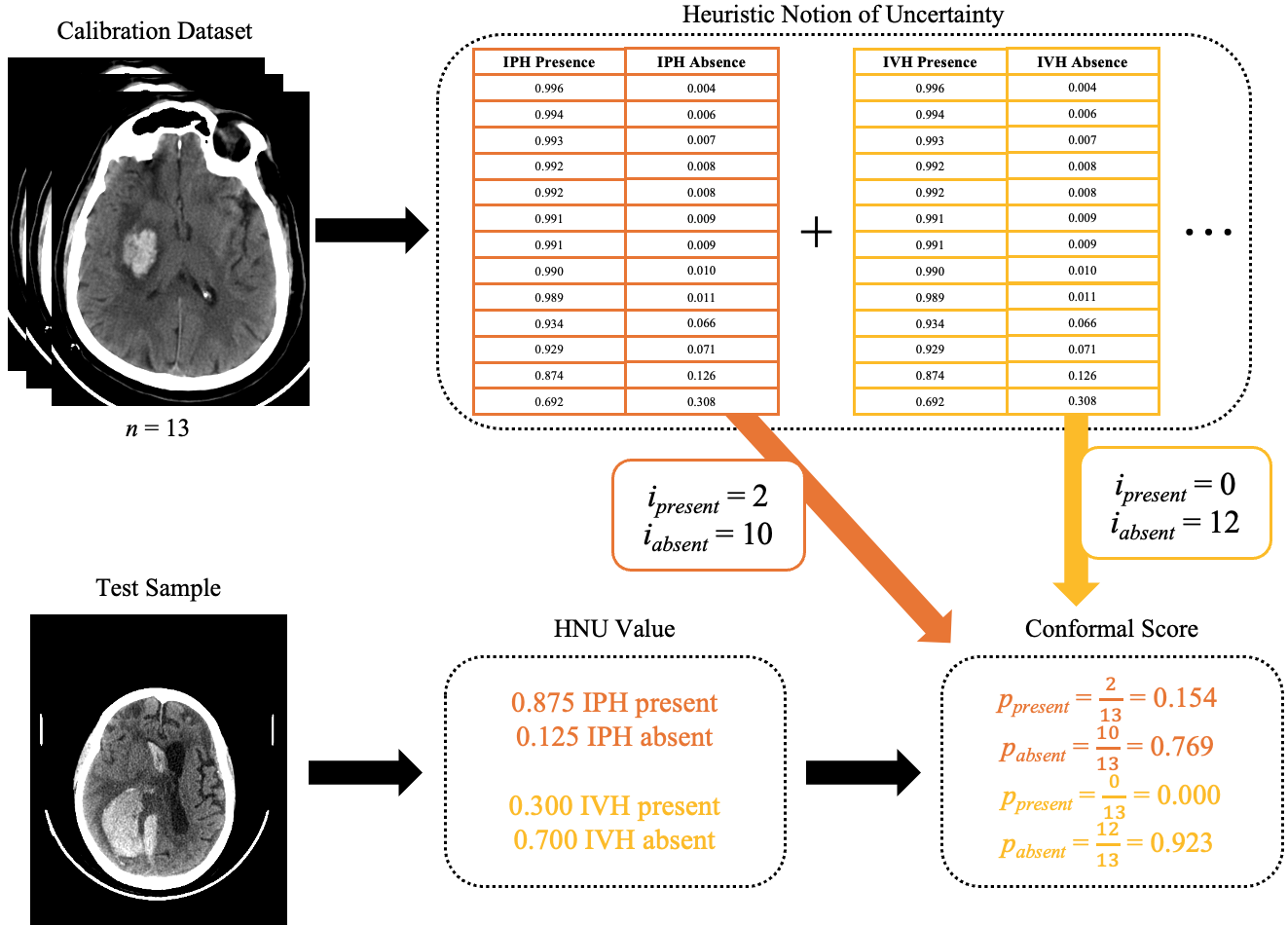}
	\caption{\centering High-level overview of Mondrian conformal prediction as it is applied in this study. IPH = intraparenchymal hemorrhage, IVH = intraventricular hemorrhage, HNU = heuristic notion of uncertainty, \textit{i}  = insertion index in sorted value array, \textit{p} = conformal score.}
	\label{fig:fig1}
\end{figure}

In addition to uncalibrated models, another source of mistrust in DL models is poor handling of out-of-domain cases. Traditional training methods restrict the apparent knowledge of a DL model to the limits of its training domain (7), such that as inputs trend away from being in-domain, the model’s predictions are increasingly likely to be incorrect. Other artifacts like patient motion or implanted devices can also significantly alter images, causing them to be out-of-domain. The out-of-domain problem is exacerbated when a model’s outputs are not calibrated.
 
In this study, we utilize MCP as a method to improve trustworthiness in DL models. We train an object detection algorithm to classify and localize ICH in non-contrast head CT slices. We demonstrate that MCP achieves state-of-the-art performance in detection of five classes of ICH, and we further show that an MCP-augmented pipeline enables highly accurate identification of challenging cases. We define challenging cases as those in which radiologists disagreed on whether ICH was present. We propose that if a DL tool can generate statistically guaranteed differential diagnoses and reliably identify when it encounters a challenging input, then physicians will be increasingly able to trust this tool and therefore better positioned to adopt it as an automated solution in their clinical or research workflow.

\section{Materials and Methods}
\subsection{Data Collection}
This is a retrospective study of 491 anonymous, non-contrast head CT scans collected at various centers in New Delhi, India and compiled as the publicly available CQ500 dataset (8).  The multi-institutional 2019 Radiological Society of North America (RSNA) Brain CT Hemorrhage Challenge dataset was used to externally validate the performance of the DL model (9). See Supplementary Descriptions 1 and 2 for more information about the datasets used in this study.

\subsection{Data Organization}
Out of the CQ500 dataset, scans for which all three readers indicated that none of the five hemorrhage classes were present were excluded and used as an evaluative control group. After narrowing the dataset to contain only scans positive for at least one class of hemorrhage, the scans were split into definite and challenging subsets. The definite dataset was defined to contain scans on whose labels readers unanimously agreed. Any scan for which there was not unanimity among the readers for the five hemorrhage classes was placed into the challenging dataset. The definite dataset was then further divided into training, tuning, calibration, and test subsets with a 70/10/10/10 split, assuring that images were grouped by their series and patient identifiers to prevent information leakage between subsets (10). Slices were also stratified by their ground truth labels for all five hemorrhage types to maintain a class distribution as close to the true distribution as possible. The data inclusion, exclusion, and splitting process is depicted in Supplementary Figure 1.
 
The external validation dataset was split into positive (hemorrhagic) and negative (non-hemorrhagic) subsets. Because the negative subset was significantly larger than the positive subset, and because there already existed a negative control group from the CQ500 dataset, a random sample of the negative subset was used during external validation. The sample size was equal to the total number of positive slices.

See Supplementary Notes 1 and 2 for additional remarks about the data organization process.

\subsection{Image Processing}
Windowing was applied to all scans in the final datasets at a window level of 50 and a window width of 80 to enhance hemorrhage prominence (11). The windowed scans were then converted from their original Digital Imaging and Communications in Medicine (DICOM) format into two-dimensional Portable Network Graphics (PNGs) of 512x512 pixels. 21,653 out of 674,254 images in the RSNA dataset could not be converted from DICOM to PNG format, so these images were excluded from the final external validation dataset. During training, the DL pipeline applied standard augmentations to reduce the effects of overfitting (12). Augmentations were not applied during other phases of the study. Additional information regarding the augmentations used can be found in Supplementary Table 1.

\subsection{Model Development}
Ultralytics’ You Only Look Once (YOLO) v8 computer vision model was selected as the DL tool for this task because it is well-regarded as state-of-the-art in object detection (13). The YOLO pipeline outputs both the model’s confidence score for a bounding box containing an object of interest, as well as its confidence score for a detected object belonging to the given class. The final confidence score for a given prediction is determined by a non-linear combination of these two confidence values whose weighting is optimized as an auxiliary component of the cost function during training. The model was trained for 300 epochs with a batch size of six using the AdamW optimization algorithm with an initial learning rate of 0.01 and YOLO’s cosine annealing learning rate scheduler (14). At the end of each training epoch, the model was applied to the tuning dataset to benchmark its performance. Additional training parameters are listed in Supplementary Table 2.

\subsection{Model Calibration}
When the model completed training, inference was performed on the calibration dataset. The model was allowed to predict up to 10,000 instances of hemorrhage per slice because to populate class-wise Mondrian groups, a bounding box and confidence score for each hemorrhage class was needed for every slice. These predictions were then filtered by identifying the highest confidence bounding box for each hemorrhage class, resulting in five confidence scores—one for each hemorrhage class—for each calibration sample. Five Mondrian groups were then formed by sorting the calibration confidence scores across all samples for each hemorrhage class. An additional five groups were formed for the absence of each hemorrhage class by sorting the additive complements of all calibration confidence scores within each hemorrhage class. Figure 1 demonstrates the flow of MCP from calibration to inference.

\subsection{Model Performance Evaluation}
After generating Mondrian groups, the model was evaluated on the test, challenging, negative control, and RSNA external validation datasets. During inference for each dataset, the model was once again allowed to predict up to 10,000 instances of hemorrhage per slice (Figure 2, Step 1). The highest confidence bounding boxes for each hemorrhage class were then selected, yielding five bounding boxes—each of a different class label—for each sample (Figure 2, Step 2). Next, clusters were created by calculating the intersection over union (IoU) between each bounding box and all other bounding boxes (Figure 2, Step 3). A cluster was defined by any bounding box whose IoU with all other bounding boxes was less than the given threshold. Then, each cluster was filled by reinspecting all initial predictions, excluding those that had already been selected, and including any bounding box whose IoU with the cluster-delimiting bounding box was greater than or equal to the given IoU threshold (Figure 2, Step 4). Penultimately, each cluster was condensed to contain only five bounding boxes by retaining the cluster-delimiting bounding box and the highest confidence bounding box for the remaining four hemorrhage classes (Figure 2, Step 5). Finally, conformal scores were calculated for each bounding box, and conformalized prediction sets were generated for each cluster (Figure 2, Step 6).

\begin{figure}[h!]
	\centering
	\includegraphics[scale=0.3]{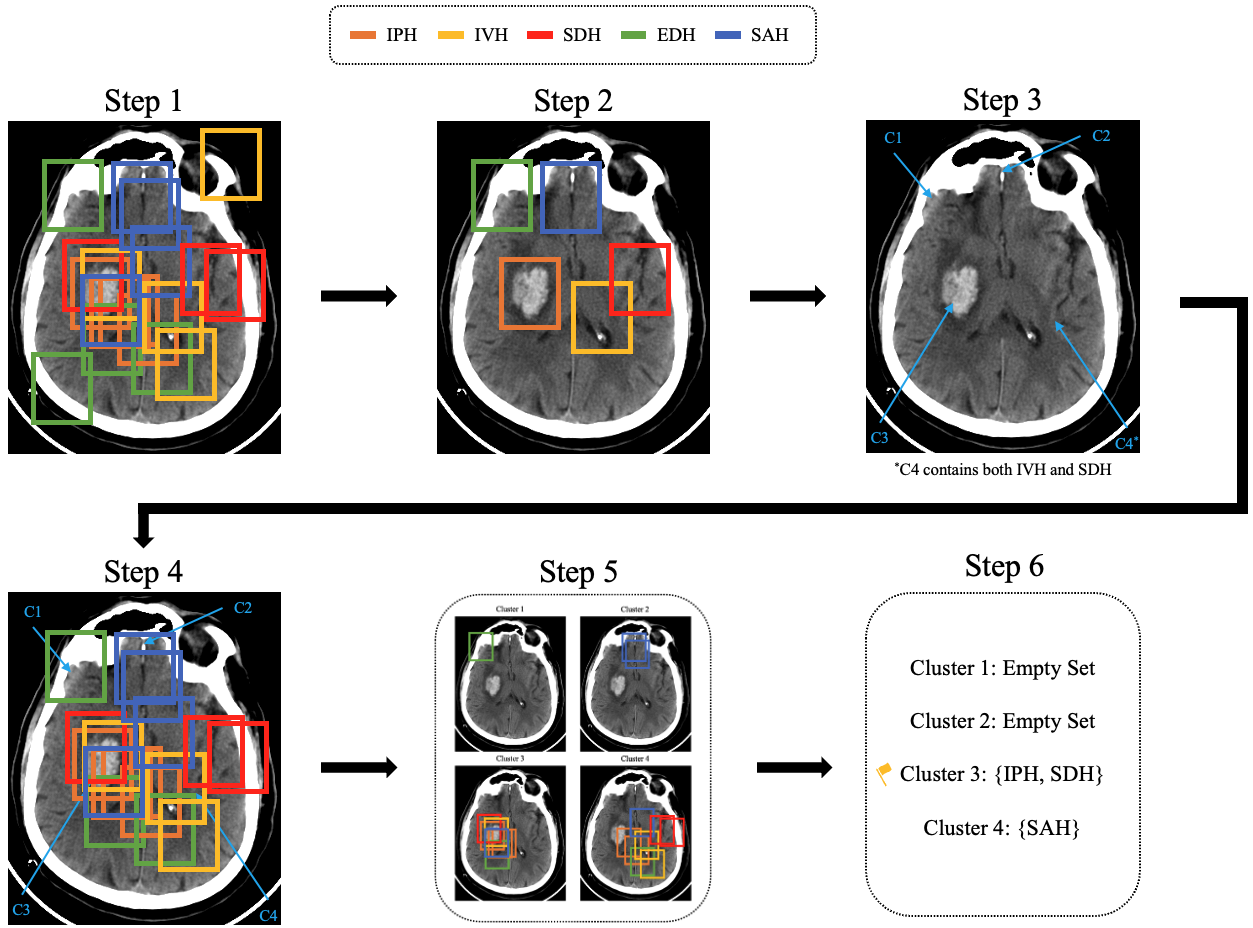}
	\vspace{-10pt}
	\caption{\centering Algorithm for clustering predictions with IoU thresholding and non-maximum suppression. IPH = intraparenchymal hemorrhage, IVH = intraventricular hemorrhage, SDH = subdural hemorrhage, EDH = epidural hemorrhage, SAH = subarachnoid hemorrhage, C1 = Cluster 1, C2 = Cluster 2, C3 = Cluster 3, C4 = Cluster 4, IoU = intersection over union.}
	\label{fig:fig1}
\end{figure}

Because this algorithm relied on two threshold values—one for the IoU and one for the conformal score—an optimization process was conducted with the test dataset to determine which thresholds to apply during evaluation on the challenging, negative, and external validation datasets. Three confusion matrices were defined. Each matrix tested different criteria for a prediction to be counted as a true positive (TP), false positive (FP), true negative (TN), or false negative (FN). Precise textual definitions of TP, FP, TN, and FN for each confusion matrix can be found in Figure 3, and visual examples of each case are depicted in Figure 4.

\begin{figure}[h!]
	\centering
	\includegraphics[scale=0.4]{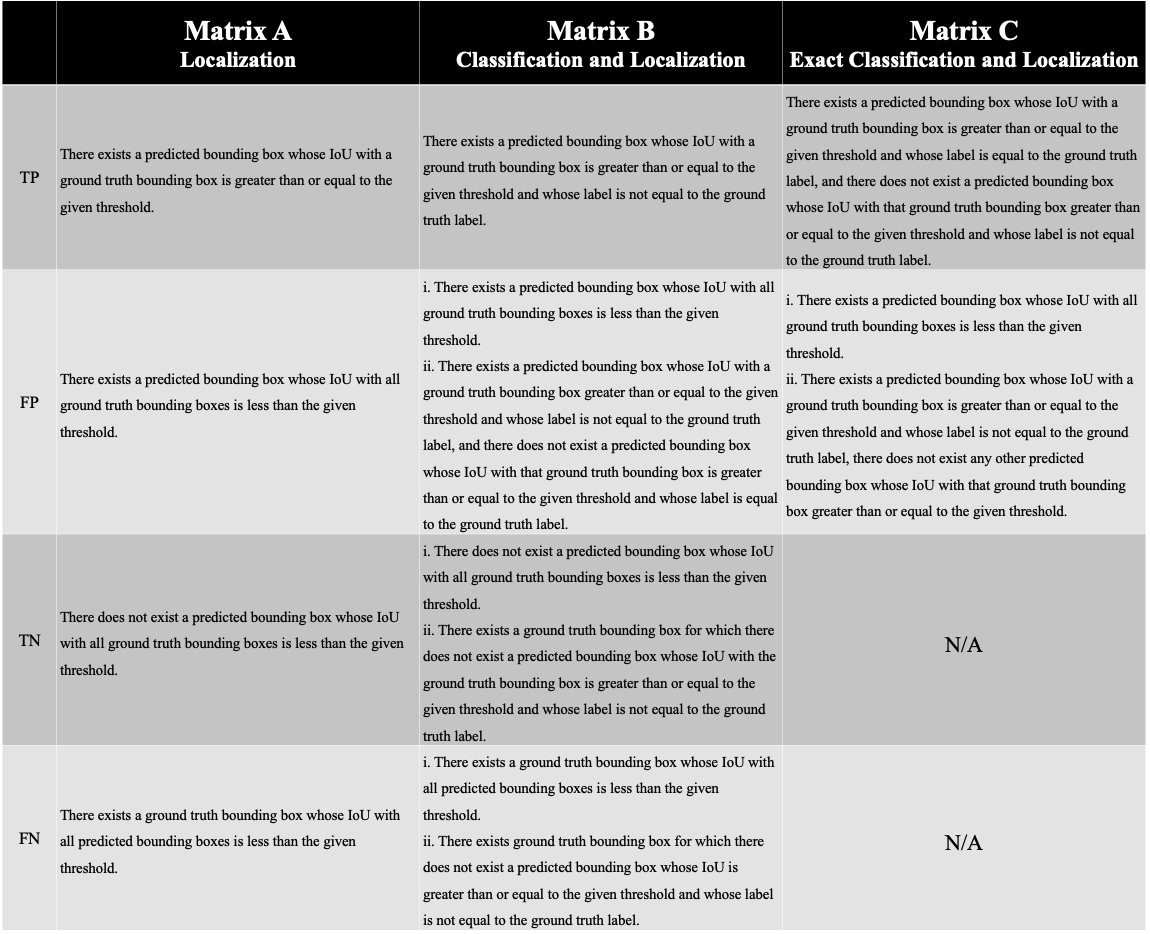}
	\vspace{-10pt}
	\caption{\centering Textual descriptions of TP, FP, TN, and FN for the three confusion matrices used during model evaluation. TP = true positive, FP = false positive, TN = true negative, FN = false negative, IoU = intersection over union.}
	\label{fig:fig1}
\end{figure}

\begin{figure}[h!]
	\centering
	\includegraphics[scale=0.31]{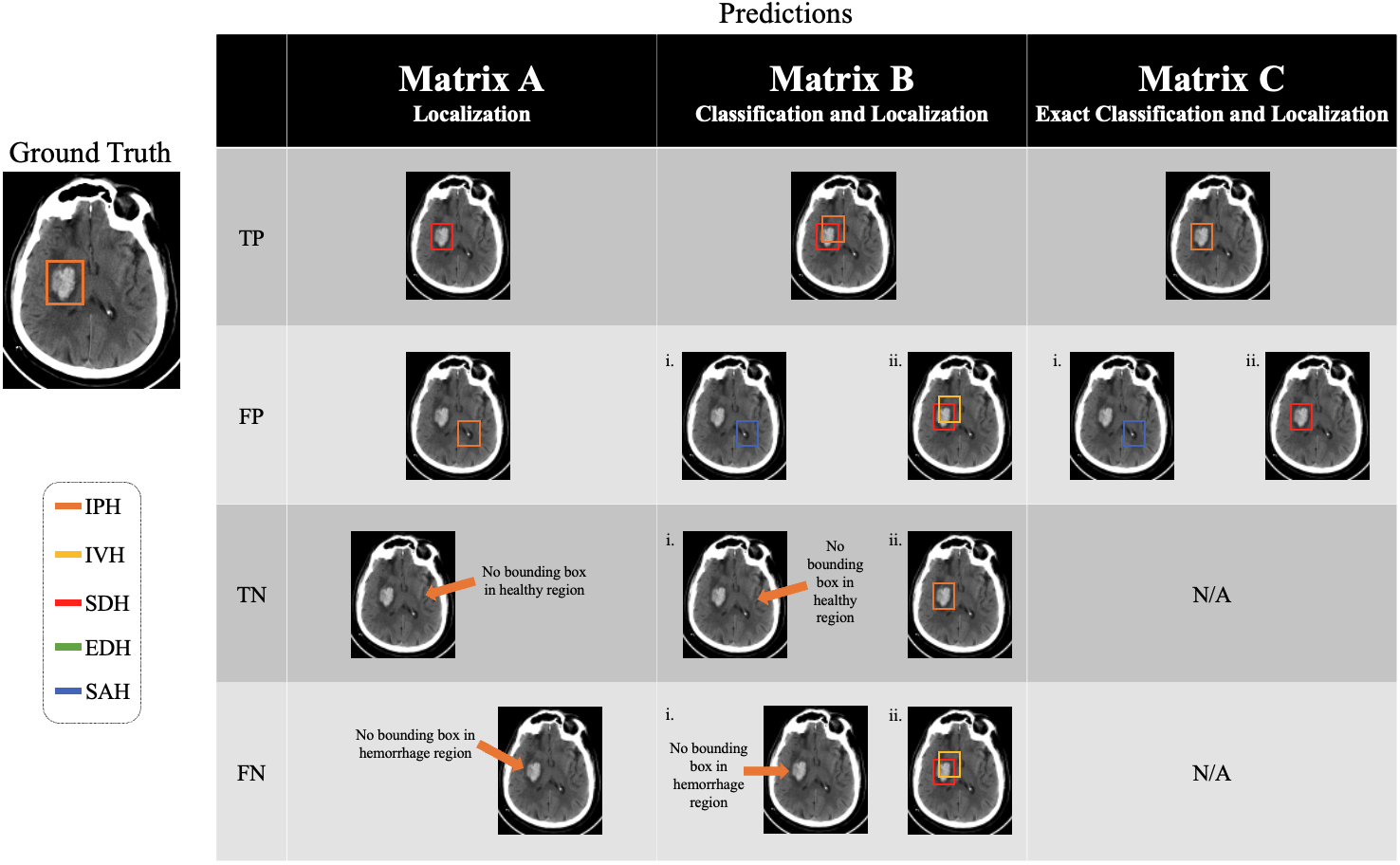}
	\vspace{-10pt}
	\caption{\centering Visual depictions of examples of TP, FP, TN, and FN for the three confusion matrices used during model evaluation. TP = true positive, FP = false positive, TN = true negative, FN = false negative, IPH = intraparenchymal hemorrhage, IVH = intraventricular hemorrhage, SDH = subdural hemorrhage, EDH = epidural hemorrhage, SAH = subarachnoid hemorrhage.}
	\label{fig:fig1}
\end{figure}
During optimization, IoU and conformal score thresholds were varied between 0.00 and 1.00 at intervals of 0.05. The clustering algorithm described above and depicted in Figure 2 was then applied using each combination of IoU and Conformal score thresholds, yielding 441 conformalized prediction sets (differential diagnoses) for each test instance. Then, for Matrices A and B, the IoU and Conformal score thresholds which yielded the maximum the area under the receiver operating characteristic curve (AUROCC) were selected. For Matrix C, because only TP and FP were defined, the threshold values which yielded the maximum positive predictive value (PPV) were selected. To avoid confusion in the Results section, note that the threshold values obtained during optimization for both Matrix A and Matrix B were identical.
 
These threshold values were then applied to the challenging, negative, and external validation datasets. For the challenging dataset, all four counts (TP, FP, TN, FN), as well as sensitivity, specificity, PPV, negative predictive value (NPV), and the F1-score were reported for the thresholds obtained from Matrices A and B. For Matrix C, TN and FN were not well-defined, so only PPV was reported. For the control group, only the TN and FP counts and specificity were reported. Other metrics were not defined because this dataset only contained non-hemorrhagic slices. No bounding boxes were available for the external validation dataset, so the evaluation was agnostic to localization performance and assessed only the model’s classification ability. As such, the definitions of TP, FP, TN, and FN were adjusted to yield meaningful results (see Figure 5). For classification, sensitivity, specificity, PPV, NPV, and the F1-score were reported. For exact classification, because FN was not well-defined, only specificity and PPV were reported.

\begin{figure}[h!]
	\centering
	\includegraphics[scale=0.3]{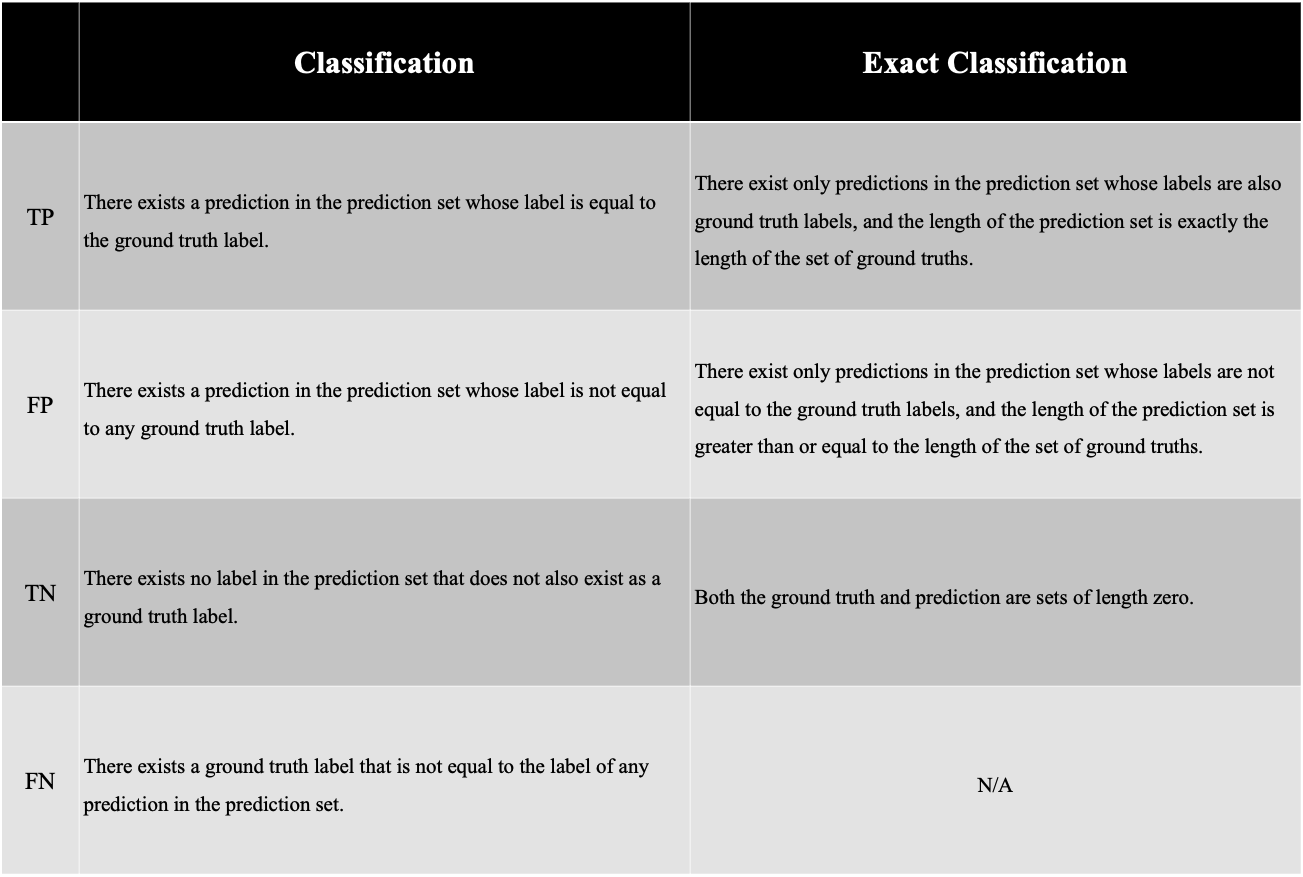}
	\vspace{-10pt}
	\caption{\centering Textual descriptions of TP, FP, TN, and FN for the two confusion matrices used during external validation of the model. TP = true positive, FP = false positive, TN = true negative, FN = false negative.}
	\label{fig:fig1}
\end{figure}
\vspace{90pt}

In addition to the metrics described above, mean average precision (mAP) was reported for all datasets, except the control group because it contained no positive samples and the external validation dataset because bounding boxes were unavailable. mAP is a standard metric in object detection and though it does not paint a full picture of a model’s performance, it can offer insight into its positive predictive ability (14,15).

\subsection{Trustworthiness Evaluation}
This study also evaluated the performance of MCP as a mechanism for trustworthiness in DL. The accuracy of the model in identifying challenging cases was reported, where a sample was predicted as challenging if the final prediction set for any cluster within the sample contained either of the following: both the presence and absence of one hemorrhage class, or the presence of more than one hemorrhage class. The number of samples in the definite test dataset which the model incorrectly labeled challenging was also reported to assess the model’s ability to distinguish between definite and challenging cases.

\subsection{Statistical Analysis}
All calculations were performed using the NumPy (version 1.24.4) and scikit-learn (version 1.2.2) libraries in Python (version 3.8.16).

\section{Results}
\subsection{Dataset Characteristics}
The definite dataset was split such that there were 10,815 training slices, 1,546 tuning slices, 1,546 calibration slices, and 1,545 test slices. The challenging dataset comprised 6,856 slices. Instance frequencies for each hemorrhage class within each dataset are shown in Table 1. 
\begin{table}[h!]
	\centering
	\begin{tabular}{c|ccccc}
	Class    & Training & Tuning & Calibration & Test  & Challenging \\ \hline
	\rule{0pt}{2ex}
	IPH      & 4,743    & 664    & 664         & 683   & 732         \\
	IVH      & 1,415    & 219    & 208         & 187   & 407         \\
	SDH      & 3,183    & 432    & 466         & 451   & 4,251       \\
	EDH      & 273      & 40     & 40          & 39    & 174         \\
	SAH      & 5,237    & 754    & 765         & 745   & 3,557       \\
	Combined & 14,581   & 2,109  & 2,143       & 2,105 & 9,121       \\ \hline
	\end{tabular} \\ 
	\vspace{6pt}
	\normalsize{Table 1: Frequency of all hemorrhage classes for each dataset. IPH = intraparenchymal hemorrhage, IVH = intraventricular hemorrhage, SDH = subdural hemorrhage, EDH = epidural hemorrhage, SAH = subarachnoid hemorrhage.}
\end{table} 
Of the 652,601 slices from the RSNA dataset successfully converted to PNGs, there were 32,564 instances of intraparenchymal hemorrhage (IPH), 23,766 instances of intraventricular hemorrhage (IVH), 42,496 instances of subdural hemorrhage (SDH), 2,761 instances of epidural hemorrhage (EDH), and 32,122 instances of subarachnoid hemorrhage (SAH). 

\subsection{Model Development Cost}
Training the model for 300 epochs required 10.867 hours on three NVIDIA A100-SXM4-80GB cards. Inference required approximately 40 additional hours on one of these cards.

\subsection{Model Performance: Hemorrhage Detection}
mAP values at an IoU threshold of 0.95 (mAP@0.95) for each hemorrhage class are displayed for each dataset in Table 2.
\begin{table}[h!]
	\centering
	\begin{tabular}{c|ccccc}
	Class    & Training & Tuning & Calibration & Test  & Challenging \\ \hline
	\rule{0pt}{2ex}
	IPH      & 0.979    & 0.861  & 0.876       & 0.873 & 0.0492      \\
	IVH      & 0.958    & 0.804  & 0.798       & 0.851 & 0.040       \\
	SDH      & 0.976    & 0.891  & 0.880       & 0.884 & 0.0538      \\
	EDH      & 0.964    & 0.889  & 0.877       & 0.867 & 0.0379      \\
	SAH      & 0.980    & 0.837  & 0.842       & 0.864 & 0.00659     \\
	Combined & 0.972    & 0.856  & 0.855       & 0.868 & 0.0375      \\ \hline
	\end{tabular} \\
	\vspace{6pt}
	\normalsize{Table 2: mAP values at an IoU threshold of 0.95 (mAP@0.95) of all hemorrhage classes for each dataset. IPH = intraparenchymal hemorrhage, IVH = intraventricular hemorrhage, SDH = subdural hemorrhage, EDH = epidural hemorrhage, SAH = subarachnoid hemorrhage.}
\end{table}
For the test and challenging datasets, confusion matrices with IoU and Conformal score thresholds optimized on the test dataset are shown in Table 3.
\begin{table}[h!]
	\centering
	\begin{tabular}{c|cccc}
	Dataset                      & Metric      & Matrix A & Matrix B & Matrix C \\ \hline
	\rule{0pt}{2ex}
	\multirow{9}{*}{Test}        & TP          & 1,952    & 1,950    & 347      \\
								 & FP          & 187      & 187      & 30       \\
								 & TN          & 1,358    & 1,358    & —        \\
								 & FN          & 153      & 155      & —        \\
								 & Sensitivity & 0.927    & 0.926    & —        \\
								 & Specificity & 0.879    & 0.879    & —        \\
								 & PPV         & 0.913    & 0.912    & 0.920    \\
								 & NPV         & 0.899    & 0.898    & —        \\
								 & F1          & 0.920    & 0.919    & —        \\ \hline
								 \rule{0pt}{2ex}
	\multirow{9}{*}{Challenging} & TP          & 1,325    & 1,325    & 15       \\
								 & FP          & 2,222    & 2,222    & 107      \\
								 & TN          & 4,634    & 4,634    & —        \\
								 & FN          & 7,796    & 7,796    & —        \\
								 & Sensitivity & 0.145    & 0.145    & —        \\
								 & Specificity & 0.676    & 0.676    & —        \\
								 & PPV         & 0.374    & 0.374    & 0.123    \\
								 & NPV         & 0.323    & 0.323    & —        \\
								 & F1          & 0.209    & 0.209    & —        \\ \hline
	\end{tabular} \\
	\vspace{6pt}
	\normalsize{Table 3: Confusion matrices and derived metrics for the test and challenging datasets. Test metrics were calculated by optimizing the IoU and Conformal score thresholds for AUROCC (Matrices A and B) and (Matrix C). Challenging metrics were calculated using the optimal thresholds as determined by the test dataset. TP = true positive, FP = false positive, TN = true negative, FN = false negative, PPV = positive predictive value, NPV = negative predictive value, IoU = intersection over union, AUROCC = area under the receiver operating characteristic curve.}
\end{table}
Table 4 depicts the model’s performance on the negative and external validation datasets using the same optimal threshold values.
\begin{table}[h!]
	\centering
	\begin{tabular}{c|ccc}
	Dataset                                                                                                         & Metric      & Matrices A and   B Thresholds & Matrix C   Thresholds \\ \hline
	\rule{0pt}{2ex}
	\multirow{3}{*}{Negative Control}                                                                             & TN          & 61,637                        & 82,074                \\
																													& FP          & 20,684                        & 247                   \\
																													& Specificity & 0.749                         & 0.997                 \\ \hline
																													\rule{0pt}{2ex}
	\multirow{9}{*}{\begin{tabular}[c]{@{}c@{}}External Validation\\ (Classification)\end{tabular}}         & TP          & 131,994                       & 94,225                \\
																													& FP          & 60,088                        & 839                   \\
																													& TN          & 127,748                       & 186,997               \\
																													& FN          & 55,842                        & 93,611                \\
																													& Sensitivity & 0.703                         & 0.502                 \\
																													& Specificity & 0.680                         & 0.995                 \\
																													& PPV         & 0.687                         & 0.991                 \\
																													& NPV         & 0.696                         & 0.666                 \\
																													& F1          & 0.695                         & 0.666                 \\ \hline
																													\rule{0pt}{2ex}
	\multirow{5}{*}{\begin{tabular}[c]{@{}c@{}}External Validation\\ (Exact Classification)\end{tabular}} & TP          & 27,295                        & 302                   \\
																													& FP          & 43,934                        & 668                   \\
																													& TN          & 60,284                        & 93,624                \\
																													& Specificity & 0.578                         & 0.993                 \\
																													& PPV         & 0.383                         & 0.311                 \\ \hline
	\end{tabular} \\
	\vspace{6pt}
	\normalsize{Table 4: Performance metrics for the negative control and external validation datasets. Metrics are reported for two sets of threshold values: the first set was determined by optimizing for AUROCC (Matrices A and B), and the second set was determined by optimizing for PPV (Matrix C). TP = true positive, FP = false positive, TN = true negative, FN = false negative, PPV = positive predictive value, NPV = negative predictive value, AUROCC = area under the receiver operating characteristic curve.}
\end{table}
\newpage

\subsection{Model Performance: Identification of Challenging Cases}
The model achieved an identification accuracy of 99.7\% (6,837 out of 6,856) at the optimized IoU and conformal score thresholds. The model labeled zero of the 1,545 definite test dataset samples as challenging. This finding is significant because the model performs markedly worse (p < 0.00001) on challenging cases than definite ones, as shown in Table 2. This demonstrates that the MCP pipeline can reduce the rate of false predictions by flagging challenging cases and referring them to a human reviewer. An example of the model’s failure on a challenging case is shown in Figure 6.
\begin{figure}[h!]
	\centering
	\includegraphics[scale=0.8]{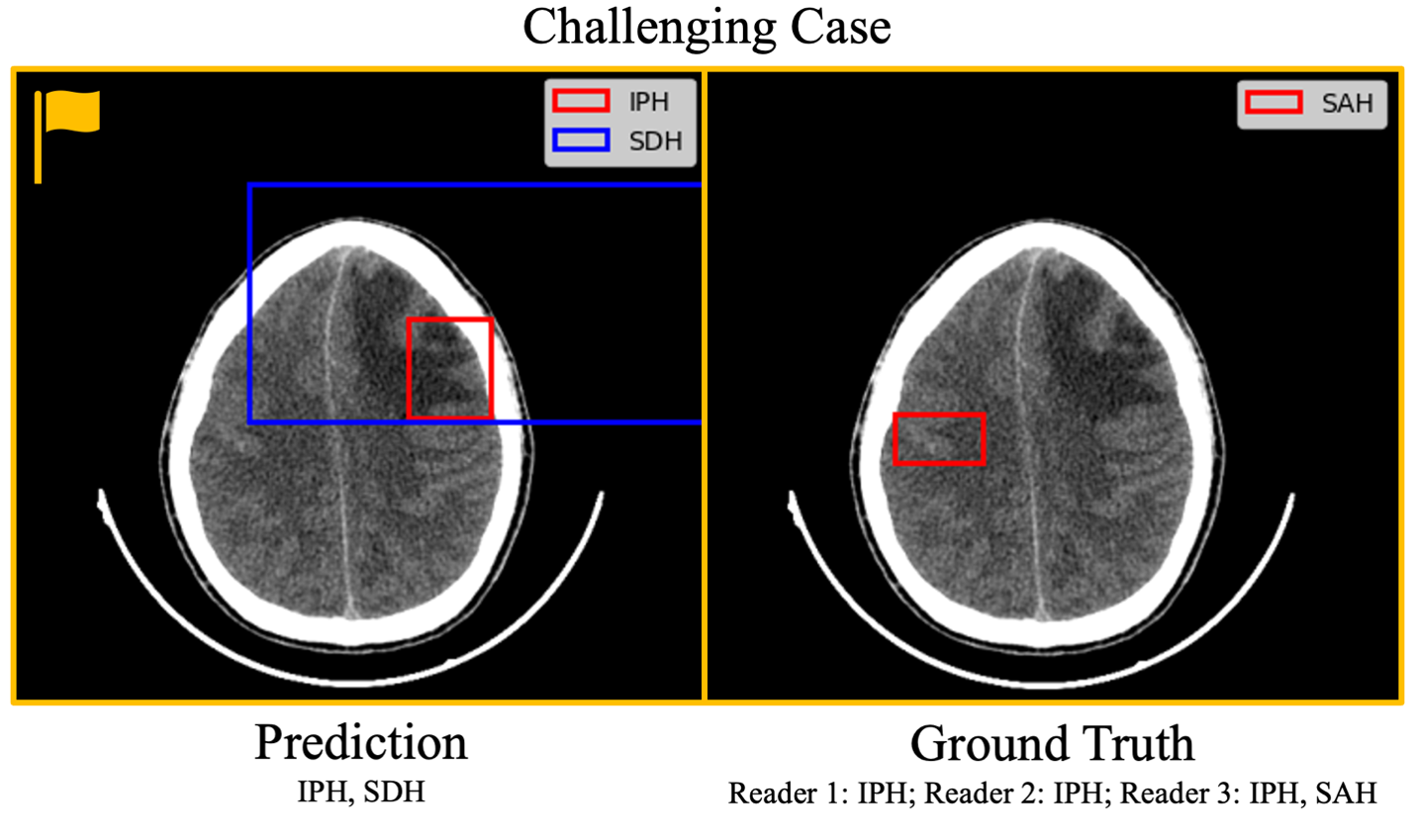}
	\vspace{-10pt}
	\caption{\centering Example of challenging case prediction and ground truth. IPH = intraparenchymal hemorrhage, SDH = subdural hemorrhage, SAH = subarachnoid hemorrhage.}
	\label{fig:fig1}
\end{figure}
\newpage
\section{Discussion}
This study aims to propose a method for increasing trust in medical DL tools by demonstrating the efficacy of MCP in the task of ICH detection and supplying a deployable toolkit for applications in other tasks. The high accuracy of the proposed approach in identifying challenging cases indicates that MCP is a promising method for enhancing trust in radiological AI systems. Furthermore, the GitHub repository and usage tutorial released to accompany this work offers a practical, scalable implementation of MCP to further validate its performance.
 
DL-assisted ICH detection was first proposed in 2017 with Phong, et al. demonstrating that a convolutional neural network could be trained to classify the presence or absence of ICH with 98.9\% average accuracy (16). Recently, Teneggi, et al. assessed the performance differences between strongly- and weakly- supervised models in ICH classification using the RSNA dataset as its primary dataset and the CQ500 dataset as external validation (17). They reported an AUROCC of 0.96 on their tuning subset of the RSNA dataset and of 0.95 on the CQ500 dataset. Additionally, Umapathy, et al. obtained an F1 score of 0.97 on the CQ500 dataset with a convolutional ensemble trained on the RSNA dataset (18). As for conformal prediction, Vovk, et al. introduced the underlying theory in 2005 (19). More recently, Angelopoulos and Bates released a comprehensive guide to conformal methods in modern DL (5), and Ding, et al. proposed class-conditional conformal prediction (6), from which the methodologies used in this study are derived.
 
This work presents a DL solution to ICH detection that performs on par with state-of-the-art models and introduces MCP as a method of building trust with radiological DL tools. The uncertainty-aware localization and classification model generates statistically guaranteed differential diagnoses, and the integration of MCP enables it to flag challenging cases with near-perfect accuracy for review by radiologists. Moreover, the DL architecture used in this solution makes it easily adaptable to a wide array of radiological and medical tasks, especially considering that the IoU and conformal score thresholds can be optimized uniquely for each task to which this pipeline is applied. This opens the door for further investigation into the critical frontier of trust in DL. The publicized code and deployment instructions further encourage the continued research and development of MCP as a mechanism for trust in radiological DL. 
 
The DL model operates on two-dimensional CT slices, so it is not immediately compatible with current Picture Archiving and Communication Systems (PACS). Given the current state of radiological DL, however, this improvement is certainly feasible without significant modification to the MCP backbone of the trust component.
 
To deem a DL algorithm trustworthy, it must be able to reliably communicate when it encounters cases about which it is uncertain. This study not only demonstrates that an MCP-based approach can yield up to 99.7\% accuracy in the identification of challenging cases, but also that a trustworthy DL model achieves performance metrics on par with state-of-the-art models. Further efforts in trustworthy radiological DL will include increasing practicality by expanding the application of MCP to three-dimensional models, as well as continuous external validation to account for generalizability. Trustworthy DL is the necessary next step toward the feasible clinical adoption of AI tools in radiology.

\newpage
\section{Supplementary Materials}
\textit{Supplementary Description 1: CQ500 Dataset} \\
The Centre for Advanced Research in Imaging, Neurosciences, and Genomics is responsible for the compilation of the CQ500 dataset. Scans were collected in two batches, B1 and B2. B1 comprised all head CTs conducted at a subset of the collection centers from 20 November 2017 to 20 December 2017. B2 scans were selectively included from a much larger dataset of scans obtained from the entire set of collection centers over an unspecified date range. First, radiology reports for scans in the larger dataset were processed with a natural language processing algorithm to identify scans with IPH, IVH, SDH, EDH, SAH, and calvarial fracture. Scans with reports containing mention of any of these conditions were then randomly selected such that there were approximately 80 scans for each intracranial condition. Scans were excluded if they contained post-operative defects, did not contain a non-contrast series with axial cuts and a soft reconstruction kernel covering the complete brain, or the patient was less than seven years of age. The final CQ500 dataset was read by three senior radiologists with eight, 12, and 20 years of experience in cranial CT interpretation. Each reader provided labels for the presence or absence of IPH, IVH, SDH, EDH, and SAH, as well as positive or negative indicators for chronic SDH, calvarial fracture, midline shift, and mass effect (8). \\ \\

\textit{Supplementary Description 2: RSNA External Validation Dataset} \\
The RSNA dataset is a multi-institutional effort with images contributed by Stanford University, Universidade Federal de São Paulo, and Thomas Jefferson University. Scans were labeled by a team of sixty junior and senior radiologists who were members of the American Society of Neuroradiology. The RSNA dataset was split between two stages for Kaggle competition; this study incorporates the 674,254 training slices from Stage 1. Annotation was performed on a commercial, browser-based annotation platform called MD.ai. Readers used their personal computers to access the software, and each reader was blind to all other readers’ annotations. Users were directed to go through a series of axial brain CT images and choose from five hemorrhage subtype image labels: IPH, IVH, SDH, EDH, and SAH. They could also label the initial and final slices of a specific hemorrhage and let the software interpolate the labels for the slices in between. In cases where multiple hemorrhage subtypes were found, multiple labels were assigned to the appropriate slices. If no hemorrhagic features were present in the entire examination, the annotator could assign one of two labels: normal or abnormal/not hemorrhage (such as stroke, atrophy, white matter disease, hydrocephalus, tumor). An extra flag related to the examination was provided to indicate incomplete or incorrectly identified body parts (9). Annotators underwent onboarding sessions in groups of 10. Practice cases were utilized to acquaint readers with both the tool and the annotation procedure. Each volunteer was given access to 24 training tests, which they labeled using the annotation tool. Ratings for both examination and image-based annotations were documented for every annotator. \\ \\

\textit{Supplementary Note 1: Ambiguous Labeling} \\
There were two small subsets of scans with ambiguous labeling: one subset contained scans which readers identified as containing hemorrhage but for which no bounding boxes were provided in the annotations, and the other subset contained scans for which bounding boxes were provided but which readers had identified as containing no hemorrhage. The first subset was excluded the from the study and the second subset was placed into the challenging dataset. \\ \\

\textit{Supplementary Note 2: Definite-Challenging Dataset Overlap} \\
There was overlap between the definite and challenging datasets because concordance was determined for each hemorrhage class, such that if for a single scan readers agreed on one hemorrhage label but disagreed on another, that scan was placed into the definite dataset with concordant label and into the challenging dataset with the disagreed-upon label. \\ \\

\newpage
\textit{Supplementary Table 1: Augmentations Applied to Training Images} \\
\begin{table}[h!]
	\centering
	\begin{tabular}{c|c}
	\textbf{Augmentation Description}                            & \textbf{Value} \\ \hline
	\rule{0pt}{2ex}
	Image Hue Saturation Value (HSV) Hue   adjustment (fraction) & 0.015          \\ 
	Image HSV-Saturation adjustment   (fraction)                 & 0.700          \\ 
	Image HSV-Value adjustment (fraction)                        & 0.400          \\ 
	Image rotation (+/- degrees)                                 & 0.000          \\ 
	Image translation (+/- fraction)                             & 0.100          \\ 
	Image scale (+/- gain)                                       & 0.500          \\ 
	Image shear (+/- degrees)                                    & 0.000          \\ 
	Image perspective (+/- fraction)                             & 0.000          \\ 
	Image flip along y-axis (probability)                        & 0.000          \\ 
	Image flip along x-axis (probability)                        & 0.500          \\ 
	Image mosaic (probability)                                   & 0.500          \\ 
	Image mix-up (probability)                                   & 0.000          \\ 
	\end{tabular}
\end{table}

\textit{Supplementary Table 2: YOLOv8 Training Hyperparameters} \\
\begin{table}[h!]
	\centering
	\begin{tabular}{c|c}
	\textbf{Hyperparameter}           & \textbf{Value}              \\ \hline
	\rule{0pt}{2ex}
	Seed                              & 829                         \\ 
	Learning Rate                     & 0.01                        \\ 
	Learning Rate Scheduler           & Cosine Annealing            \\ 
	Non-Max Suppression Threshold     & 0.45                        \\ 
	Confidence Threshold              & 0.25                        \\ 
	Intersection over Union Threshold & 0.50                        \\ 
	Convolutional Filters per Layer   & 64                          \\ 
	Convolutional Layers              & 53                          \\ 
	Activation Function               & Leaky Rectified Linear Unit \\ 
	Number of Anchors                 & 3                           \\ 
	Network Architecture              & YOLOv8x                     \\ 
	Optimizer                         & AdamW                       \\ 
	Cost Function                     & Varifocal Loss              \\ 
	\end{tabular}
\end{table}
\newpage
\textit{Supplementary Figure 1: Flowchart of Data Inclusion, Exclusion, and Splitting}
\begin{figure}[h!]
	\centering
	\includegraphics[scale=0.3]{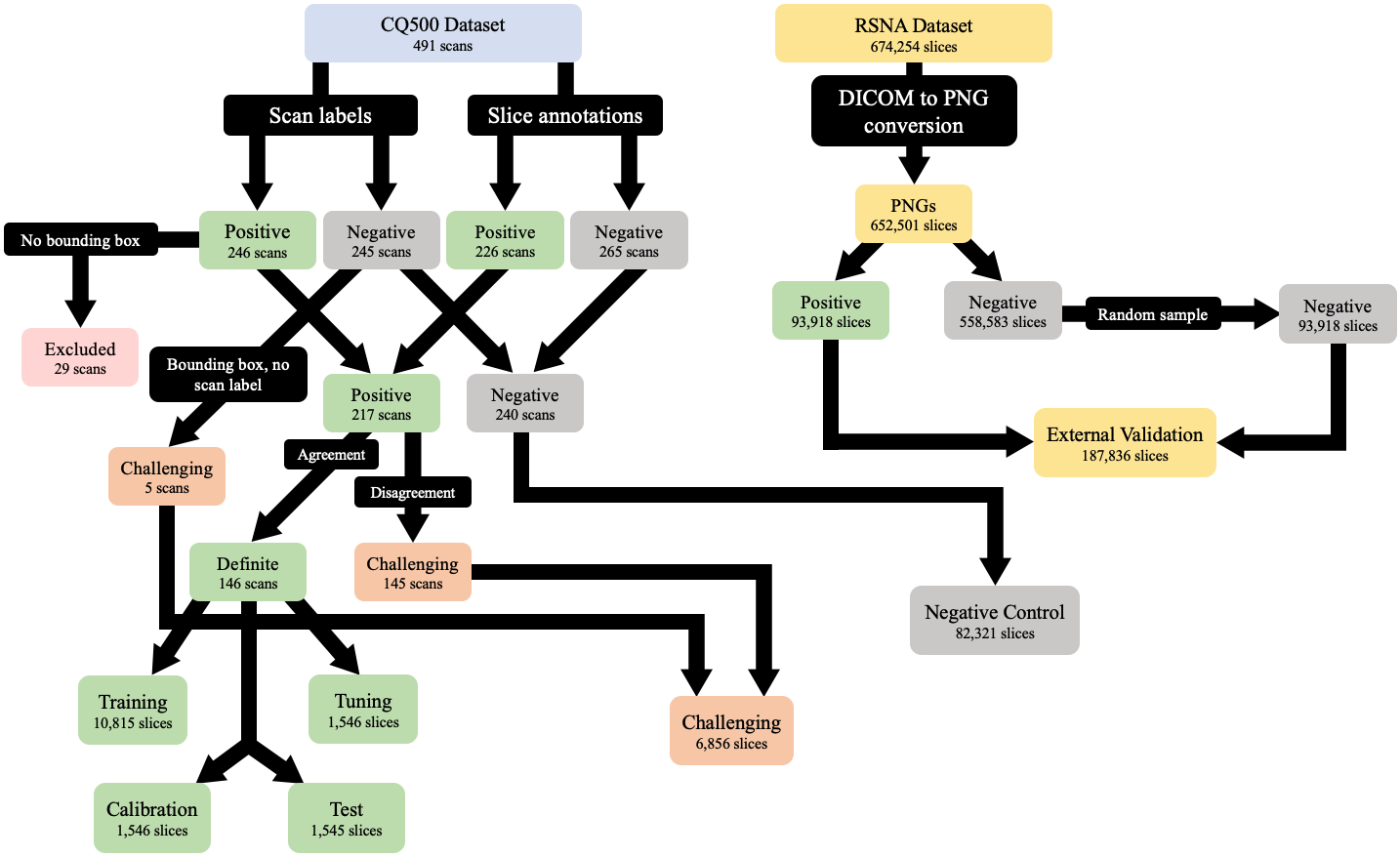}
	\vspace{-10pt}
\end{figure}\\
\begin{center}Supplementary Figure 1: Depiction of data inclusion, exclusion, and splitting. RSNA = Radiological Society of North America, DICOM = Digital Imaging and Communications in Medicine, PNG = Portable Network Graphics.\end{center}

\end{document}